\documentclass[11pt]{article}
\usepackage{colacl06}
\usepackage{times}
\usepackage{latexsym}
\usepackage{url}
\usepackage{graphicx,subfigure}
\usepackage{devanagari}
\usepackage{amsmath}
\usepackage{multirow}
\usepackage{float}
\usepackage{amssymb}
\title{Words are not Equal: Graded Weighting Model for building Composite Document Vectors}

\author{Pranjal Singh\\
  B.Tech.-M.Tech. Dual Degree\\
  Computer Science \& Engineering\\
  Indian Institute of Technology Kanpur\\
  {\tt pranjals16@gmail.com}  \And
  Amitabha Mukerjee\\
  Professor\\
  Computer Science \& Engineering\\
  Indian Institute of Technology Kanpur\\
  {\tt amit@cse.iitk.ac.in}}

\date{}

\begin{document}
\maketitle
\begin{abstract}
Despite the success of distributional semantics, composing phrases from word vectors remains an important challenge.  Several methods have been tried for benchmark tasks such as sentiment classification, including word vector averaging, matrix-vector approaches based on parsing, and on-the-fly learning of paragraph vectors. Most models usually omit stop words from the composition.  Instead of such an yes-no decision, we consider several graded schemes where words are weighted according to their discriminatory relevance with respect to its use in the document (e.g., idf). Some of these methods (particularly tf-idf) are seen to result in a significant improvement in performance over prior state of the art.  Further, combining such approaches into an ensemble based on alternate classifiers such as the RNN model, results in an 1.6\% performance improvement on the standard IMDB movie review dataset, and a 7.01\% improvement on Amazon product reviews. Since these are language free models and can be obtained in an unsupervised manner, they are of interest also for under-resourced languages such as Hindi as well and many more languages.  We demonstrate the language free aspects by showing a gain of 12\% for two review datasets over earlier results, and also release a new larger dataset for future testing~\cite{singh-15}.
\end{abstract}

\section{Introduction}
Language representation is a very crucial aspect to perform various NLP tasks and has been looked into great detail in recent times. Language representation models have fallen into broadly two categories: ones which require hand-trained language databases such as treebanks (e.g., \cite{Socher:13}) and ones which are language agnostic and work on raw corpora (e.g., LDA, BOW, SkipGram, NLM, etc.). Liu \shortcite{Liu:15} compare various language agnostic models for topic modeling.

Language independent models such as LDA  and BOW have been quite effective since long time. Variants of BOW such as tf-idf had changed the perception of researchers towards these models when they were proved effective in various NLP tasks. LDA was able to model inter and intra documental statistical and relational structure quite well overcoming the drawbacks of BOW. But the semantic and syntactical dependencies were still ignored. After the introduction of neural language vector models, NLP saw a huge diversion in representation of words and documents.
\begin{table}[h]
\centering
\small
\begin{tabular}{|p{3.7cm}|p{0.7cm}|p{0.95cm}|p{0.6cm}|}
\hline
\textbf{Method}                                                             & \textbf{IMDB}  & \textbf{Amazon} & \textbf{Hindi} \\ \hline
RNNLM (Baseline)                                               & 86.45          & 90.03           & 78.84          \\ \hline
Paragraph Vector \cite{Le:14}                                               & 92.58          & 91.30           & 74.57          \\ \hline
Averaged Vector                                                             & 88.42          & 88.52           & 79.62          \\ \hline
Weighted Average Vector                                                            & 89.56          & 88.63               & 85.90          \\ \hline
Composite Document Vector                                     & 93.91          & 92.17               & 90.30         \\ \hline
\end{tabular}
\caption {Comparison of accuracies on 3 Datasets (IMDB, Amazon Electronics Review and Hindi Movie Reviews (IITB)) for various types of document composition models. The state of the art for these tasks are: IMDB: 92.58\%~\cite{Le:14}; Amazon:85.90\%~\cite{Dredze:08}, Hindi:79.0\%~\cite{Bakliwal:12}.}
\label{table:3Datasets}
\end{table}
For individual words, vectors are obtained via distributional learning, the mechanisms for which varies from document-term matrix factorization~\cite{Landauer:97}, various forms of deep learning~\cite{Collobert:08,Turian:10,Socher:13}, optimizing models to explain co-occurrence constraints ~\cite{Mikolov:13a,Pennington:14}, etc. Once the word vectors have been assigned, similarity between words can be captured via cosine distances. The same models have been extended (\cite{Le:14}) with new variables to build vector models for sentences and documents. These models include the essence of individual words as well as their relative order in terms of sentence vector which was earlier absent in word vectors. The advantage of these approaches is that they can capture both the syntactic and the semantic similarity between words/documents in terms of their projections onto a high-dimensional vector space; further, it seems that one can tune the privileging of syntax over semantics by using local as opposed to large contexts~\cite{Huang:12}.

Some grammarians have been trying to find whether sentence meaning accrues by combining word meanings, or whether words gain their meanings based on the context they appear in \cite{Matilal:90}. 
\cite{Turney:10} give a detailed overview of various vector space models and their composition. A surprising event in Information Theory has higher information content than an expected event (Shanon, 1948). The same happens when we give weights to word vectors. We give more weight to events which evoke surprise and less weight to events which are expected. The most popular weighting concept in this domain is the idea of tf-idf which we have utilized in this work (Refer Table \ref{table:3Datasets}).

\begin{figure}[h]
\centering
\includegraphics[width=60mm, height=35mm]{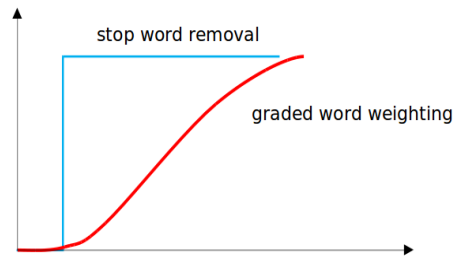}
\caption{Intuition behind the proposed approach is that graded mechanisms for word combination may do better than discrete models that simply reject stop words and treat all other words equally(e.g. averaging methods).
\label{fig:step}}
\end{figure}

In this work, we focus on a graded approach to assessing the importance of each word in a compositional models.  Graded models such as tf-idf have long been used in NLP, but they do not seem to have been used in word vector composition tasks yet.  The intuition is that in discrete cutoff functions, while simple, raise questions regarding threshold (what constitutes a stop word), and do not degrade performance gradually (Fig.~\ref{fig:step}).

We claim that the document vectors in hand is a much better representation of each document than doing it separately. This can be justified by the fact that we now incorporate contribution of each
word as per its importance as well as well as that of document without ignoring tf-idf representation which performs considerably well in tasks such as retrieval. Hence, we call this as composite document vector representation. We then go a step ahead to build an ensemble of our model and recurrent neural network, which essentially has the properties of a generative model, to achieve state-of-the-art result on IMDB movie review dataset (94.19\%) and on Amazon electronics reviews dataset with a significant improvement over previous best.
The world class results in English clearly indicate the efficacy of our approach and improvements in Hindi depict the deficiency in other models which were used earlier.

\section{Related Work}
\subsection{Neural Language Model}
Language modeling problem involved using frequency counts of n-grams for so many years but it ignores a large number of n-grams which are not seen while training leading to data sparsity and over-fitting of training data. Also these n-gram models along with BOW suffers from the curse of dimensionality. Neural Networks tend to overcome the drawbacks of n-gram models because they can model continuous variables or distributed representation, which is a necessity if we would like to find better generalizations over the highly discrete word sequences~\cite{Bengio:03}. 
Neural language models were introduced by Bengio et al., 2001 (revised in 2003\cite{Bengio:03}). They build a mapping $C$ from each word $i$ of the vocabulary $V$ to a feature vector $C(i) \in \mathbb{R}^m$, $m$ is the number of features; a probability function $g$ over words expressed with $C$; and finally learn the word vector and parameters of probability function.
Morin\shortcite{Morin:05} proposed a hierarchical model to speed up the training cost by clustering similar words before computing their probability in order to only have to do one computation per word cluster
at the output layer of the NN.
Le\shortcite{Le:11} combined neural networks with n-gram language models in a unified approach. They cluster words to structure the output vocabulary. Mikolov\shortcite{Mikolov:10} achieved the best reduction in perplexity by using recurrent neural network which uses the the current input as well as the output of the previous iteration.
Mikolov\shortcite{Mikolov:11} present several modifications of the original recurrent neural network language model (RNN LM). The present approaches that lead to more than 15 times speedup for both training and testing phases.
Collobert\shortcite{Collobert:08} show the use of semi-supervised learning using deep neural networks to perform at the state-of-the-art of various NLP tasks. Wang\shortcite{Wang:14} propose a word vector neural-network model, which takes both sentiment and semantic information into account. This word vector expression model learns word semantics and sentiment at the same time as well as fuses unsupervised contextual information and sentence level supervised labels. Neelakantan\shortcite{Neelakantan:14} took word vector models to next level where they proposed multiple embeddings per word.
The problem that still remains in hand is that either these models are computationally expensive or they have failed to generalize properly. We, therefore, adopt skipgram model~\cite{Mikolov:13a}, details about which have been discussed in next section, because deep network model of Collobert et al.(2008) takes too much time for training (skipgram reduces computational complexity from O(V) to O(log V) \cite{Morin:05}).

\subsection{Sentiment Analysis}
Majority of the existing work in this field is in English~\cite{Liu:12}. Medagoda\shortcite{Medagoda:13} surveys sentiment analysis in non-English languages while \cite{Sharma:14} give a summary of work done in Hindi in the field of opinion mining.
There have been heuristic based and machine learning based models used in this domain. Heuristic based methods, in general, classify text sentiments on the basis of total number of derived positive or negative sentiment oriented features. But these models rely heavily on human engineered features which, in general, is a domain and language dependent task. Several groups have attempted to improve the situation by modeling the composition of words into larger contexts \cite{Le:14,Socher:13,Johnson:14,Baroni:14}.

Pang\shortcite{Pang:04} achieved an accuracy of 87.2\% (Pang et al. 2004) on a dataset that discarded objective sentences and used text categorization techniques on the subjective sentences. Le\shortcite{Le:14} use paragraph vector model and obtain 92.6\% accuracy on IMDB movie review dataset.  More difficult challenges involve short texts with non-standard vocabularies,as in twitter.  Here, some authors focus on building extensive feature sets (e.g. Mohammad et al.(2013); F-score 89.14).

However, most of the work on sentiment analysis in Hindi has not attempted to form richer compositional analyses. For the type of corpora used here, the best results, obtained by combining a sentiment lexicon with hand-crafted rules (e.g. modeling negation and "but" phrases), reach an accuracy of 80\%~\cite{Mittal:13}. Joshi\shortcite{Joshi:10} compared three approaches: In-language sentiment
analysis, Machine Translation and Resource Based Sentiment Analysis. By using WordNet linking, words in English SentiWordNet were replaced by equivalent Hindi words to get H-SWN. The final accuracy achieved by them is 78.1\%. Bakliwal\shortcite{Bakliwal:12} traversed the WordNet ontology to antonyms and synonyms to identify polarity shifts in the word space. Further improvements were achieved by using a partial stemmer (there is no good stemmer / morphological analyzer for Hindi), and focusing on adjective/adverbs (seed words given to the system); their final accuracy was 79.0\% for the product review dataset. Mukherjee et al. (2012) presented the inclusion of discourse markers in a bag-of-words model and how it improved the sentiment classification accuracy by 2-4\%.

Many approaches seek to improve their performance by combining POS-tags and even parse tree structures into the models for higher accuracies in specific tasks~\cite{Socher:13}. One problem in this approach is that of  combining the word vectors to build document vectors because of issues in merging parse trees. Also these models are language dependent and computationally very expensive.

\section{Method}
The algorithms and data structures used in this thesis have been introduced and discussed below.
\subsection{Distributed Representation}
Mikolov et al. (2013b) proposed two neural network models for building word vectors from large unlabeled corpora; Continuous Bag of Words(CBOW) and Skip-Gram.  In the CBOW model, the context is the input, and one tries to learn a vector for the central word; in Skip grams, the input is the target word and one tries to guess the set of contexts. We have adopted skipgram model to build vector representations for words as it performs better with larger vocabulary.

Each current word acts as an input to a log-linear classifier with continuous projection layer, and predict words within a certain range before and after the current word. The objective is to maximize the probability of the context given a word within a language model:
\begin{center} $p(c|w;\theta)=\frac{\exp^{v_c.v_w}}{\sum_{c' \in C}\exp^{v_c.v_w}}$ \end{center}
where $v_c$ and $v_w$ $\in$ $R^d$ are vector representations for context $c$ and word $w$ respectively. $C$ is the set of all available contexts. The parameters $\theta$ are $v_{c_i}$, $v_{w_i}$ for $w \in V$, $c \in C$, $i \in 1,....,d$ (a total of $|C| \times |V| \times d$ parameters).

This distributed representation of sentences and documents~\cite{Le:14} modifies word2vec (Skip-Gram) algorithm to unsupervised learning of continuous representations for larger blocks of text, such as sentences, paragraphs or entire documents. The algorithm represents each document by a dense vector which is later trained and tuned to predict words in the document. In this framework, every paragraph is mapped to a unique vector and id, represented by a matrix $D$, which is a column matrix. Every word is mapped to a unique vector and word vectors are concatenated or averaged to predict the context, i.e., the next word.\\
The paragraph vector is shared across all contexts generated from the same paragraph but not across paragraphs. The word vector matrix W, however, is shared across paragraphs. i.e., the vector for "good" is the same for all paragraphs. The paragraph vector represents the missing information from the current context and can act as a memory of the topic of the paragraph. The advantage of using paragraph vectors is that they inherit the property of word vectors, i.e., the semantics of the words. In addition, they also take into consideration a small context around each word which is in close resemblance to the n-gram model with a large n. This property is crucial because the n-gram model preserves a lot of information of the sentence/paragraph, which includes the word order also. This model also performs better than the Bag-of-Words model which would create a very high-dimensional representation that has very poor generalization.

Our model incorporates property of document vector as well as property of word vectors to build an enhanced representation of documents without ignoring the properties of tf-idf representation.


\subsection{Semantic Composition}
\label{sec:composition}
The Principle of Compositionality is that meaning of a complex expression is determined by the meaning of its parts or constituents and the rules which guide this combination. It is also known as \emph{Frege's Principle}. In our case, the constituents are word vectors and the expression in hand is the sentence/document vector. For example,
\begin{center}
\emph{The movie is funny and the screenplay is good}
\end{center}

\begin {table}[H]
\centering
\begin{tabular}{ |c|c| }
\hline
Composition & Accuracy \\ \hline \hline
Multiplication & 50.30 \\ \hline
Average & 88.42 \\ \hline
Idf Graded Weighted Average & \textbf{89.56} \\ \hline
\end{tabular}
\caption {Results of Vector Composition with different Operations}
\label{table:composition}
\end{table}
Analyzing the results from Table \ref{table:composition}, we observed that when we deal with large number of features, there is a presence of large number of \emph{zeros} and presence of a single zero in a feature will make that features contribution zero in the final vector, which happens in our case and thus multiplicative composition fails.\\
We, therefore, adopt both simple and idf weighted average methods in our work. The advantage with addition is that, it doesnot increase the dimension of the vector and captures high level semantics with ease. In fact, \cite{Zou:13} have used simple average to construct phrase vectors which they have later used to find phrase level similarity using cosine distance.\\
\cite{Mikolov:13c} showed that relations between words are reflected to a large extent in the offsets between their vector embeddings. They also use additive composition to reflect semantic dependencies.
\begin{center}
\emph{queen - king $\approx$ woman - man}
\end{center}
\cite{Blacoe:12} clearly show that vectors of Neural Language Model and Distributed Model when used with additive composition outperform those with multiplicative composition in Paraphrase Classification task. DM vectors outperform by nearly giving accuracy difference of 6\%. They also perform very well on Phrase similarity tasks.\\
We, therefore, propose graded weighting schema for better composition of vectors which is described below.

\subsubsection{Graded Weighting}
\label{sec:graded_weighting}
We describe two approaches to incorporate graded weighting into word vectors for building document vectors. Let $v_{w_i}$ be the vector representation of the $i^{th}$ word. Then document vector $v_{d_i}$ for $i^{th}$ document is:
$$
v_{d_i} = \left\{
        \begin{array}{ll}
            0 & \quad w_k \in stopwords \\
            \sum\limits_{w_k \in d_i} v_{w_k} & \quad w_k \notin stopwords
        \end{array}
    \right.
$$
The above equation is 0-1 step-function which ignores contribution of all stop words. Now we propose another schema which weighs the contribution of each word while building document vector with a graded approach. We define $idf(t,d)=\log(\frac{|D|}{df(t)})$ where $t$ is the term, $d$ is the document and other notations are same as in previous subsection. The new document vector representation considering this graded schema is:
$$
v_{d_i} = \left\{
        \begin{array}{ll}
            0 & \quad idf(w_k,d_i) \leq \delta \\
            \sum\limits_{w_k \in d_i} idf(w_k,d_i).v_{w_k} & \quad otherwise
        \end{array}
    \right.
$$
where $\delta$ is a pre-defined threshold below which the word has no importance and above which the \emph{idf} terms gives importance to that particular word.\\
Till date, everyone has ignored how to effectively use vector composition techniques and as a result, this area has seen very less attention. But we have successfully used \emph{idf} values to give weights to word vectors and hence obtain much better sentence/document vectors. The advantage of this model is that once we obtain \emph{idf} values from training corpus, we can directly use it with test corpus without any additional computation. The results (see \ref{sec:results}) obtained by using this technique clearly demonstrate how effective it is for tasks such as sentiment analysis.

\subsection{Composite Representation}
This experiment redefined document representation in NLP used for sentiment classification. It has the property of including both syntactic and semantic properties of a piece of text. The limitations of skip-gram word vectors have been fulfilled by document vectors and hence we achieve state-of-the-art results on IMDB movie review dataset as well as amazon electronics review dataset.

We first generated $n$-dimensional word vectors by training skip-gram model on the datasets. We then assigned weights to word vectors for each document to create document vectors. This now acts as a feature set for that particular document. We then created \emph{tf-idf} vectors for each document. This can be seen as a vector representation of that particular document. We then concatenated these document vectors with document vectors obtained after training the desired dataset separately with the model proposed in \cite{Le:14}. Discrimination weighted vectors give a great boost to classification accuracies on various datasets and hence justifies our claim.

\subsection{Dimensionality Reduction}
\label{sec:dimensionality_reduction}
Dimensionality Reduction is the process of reducing the number of random variables in such a way that the remaining variables effectively reproduce most of the variability of the dataset.
The reason for using such techniques is because of the \emph{curse of dimensionality} which is a phenomena that occurs in high-dimension but doesn't occur in low-dimension.\\
Table \ref{table:700_movie_features} summarizes how feature selection has improved classification accuracy on the 700 Movie review dataset. With ANOVA-F, we selected around 4k features but with PCA, this number was just 50. So, the low accuracy with PCA can be attributed to the fact that we may have lost some important features in low dimension. Also, PCA cannot work with size of dimension $d>$\emph{size of learning set}. This sharp decrease in accuracy in both cases happens because ANOVA-F selects features with larger variance across group and thus reduces noise to a larger extent whereas PCA reduces angular variance which is not effective in this case due to the distribution of data points in high-dimensional space.
\begin{table}[H]
\small
\begin{tabular}{|p{3.6cm}|p{1.5cm}|p{1.4cm}|}
\hline
\textbf{Method} & \textbf{Feature Selection} & \textbf{Accuracy} \\ \hline
\multirow{3}{*}{Document Vector + tfidf}              & None      & 74.57 \\ \cline{2-3} 
                                              & PCA(n=50) & 76.33 \\ \cline{2-3} 
                                              & ANOVA-F & 88.07 \\ \hline
\multirow{3}{*}{Weighted Word Vector + tfidf} & None      & 76.43 \\ \cline{2-3} 
                                              & ANOVA-F   & 90.37 \\ \cline{2-3} 
                                              & PCA(n=50) & 78.61 \\ \hline
\end{tabular}
\caption {Accuracies on our newly released 700-Movie Review Dataset}
\label{table:700_movie_features}
\end{table}

This newly released dataset is much larger than previous standard dataset and very less focused towards sentiment of the review.

\section{Experiment}
\label{sec:experiment}
In this section, we describe the experiments and analyze the results.

\subsection{Datasets}
We have used 3 datasets for experiments in Hindi and 5 for English. All the datasets including our self created Hindi dataset are described below.\\
We experimented on two Hindi review datasets. One is the Product Review dataset (LTG, IIIT Hyderabad) containing 350 Positive reviews and 350 Negative reviews. The other is a Movie Review dataset (CFILT, IIT Bombay) containing 127 Positive reviews and 125 Negative reviews. Each review is around 1-2 sentences long and the sentences are mainly focused on sentiment, either positive or negative.\\
Our 700-Movie Review Corpus in Hindi contains movie reviews from websites such as \emph{Dainik Jagran} and \emph{Navbharat Times}. The movie reviews are longer than the previous corpus and contains subjects other than sentiment. There are in total 697 movie reviews from both the websites. The statistics compiled is described below.

For experiments in English, we trained on IMDB movie review dataset (Maas et al.(2013)) which consists of 25,000 positive and 25,000 negative reviews. It also contains an additional 50,000 unlabeled documents for unsupervised learning.
\begin {table}[h]
\centering
\begin{tabular}{ |l|l| }
\hline
Positive Reviews & 356 \\ 
Negative Reviews & 341 \\
Total Reviews & 697\\ \hline
\multicolumn{2}{|c|}{29.7 sentences per document} \\ \hline
\multicolumn{2}{|c|}{494.6 words per document} \\
\hline
\end{tabular}
\caption {Statistics of Movie Reviews of the 700-Movie Reviews Dataset}
\end{table}

The Trip Advisor Review dataset contains around 240K reviews (206MB) from hotel domain. Reviews with overall rating $>=$3 were annotated as positive and those with overall rating $<$3 were annotated as negative. The dataset was split into 80-20 ratio for training and testing purpose.

We also took amazon reviews for our experiments. Reviews with overall rating $>=$3 were annotated as positive and those with overall rating $<$3 were annotated as negative. The dataset was split into 80-20 ratio for training and testing purpose. There were 3 review datasets: Electronics dataset consists of 1,241,778 reviews, Watches Dataset consists of 68,356 reviews and MP3 Dataset consists of 31,000 reviews.

\subsection{SkipGram or CBOW}
\begin{figure}[ht!]
\centering
\includegraphics[width=80mm, height=60mm]{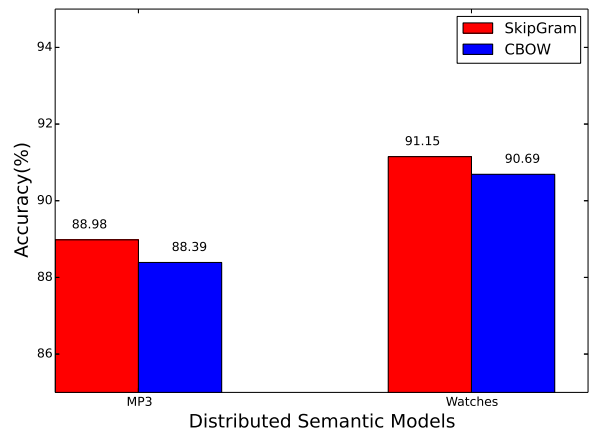}
\caption{Variation of Accuracy with skipgram and cbow on Watches and MP3 Datasets. \label{fig:accuracy_sgcbow}}
\end{figure}

We present an interesting experiment to demonstrate that skipgram indeed performs better than CBOW. SkipGram model tends to predict a context given a word whereas CBOW model predicts a word given a context. It seems intuitive and also from observation~\cite{Mikolov:13b} that SkipGram will perform better on semantic tasks and CBOW on syntactic tasks. We now try to evaluate how they differ on classification accuracies on the two datasets: \emph{Watches} and \emph{MP3}. Figure \ref{fig:accuracy_sgcbow} show that skipgram outperforms CBOW on sentiment classification task. It can be justified by the fact that sentiment inclination of a document is more oriented towards semantics of that document rather than just syntax and our results clearly demonstrate this fact.

\subsection{Results}
\label{sec:results}

\begin {table}[H]
\centering
\small
\begin{tabular}{ | c | c | }
\hline
\textbf{Method} & \textbf{Accuracy} \\ \hline
Maas et al.(2011) & 88.89\\ \hline
NBSVM-bi (Wang \& Manning, 2012) & 91.22\\ \hline
NBSVM-uni (Wang \& Manning, 2012) & 88.29\\ \hline
SVM-uni (Wang \& Manning, 2012) & 89.16\\ \hline
Paragraph Vector (Le and Mikolov(2014)) & 92.58\\ \hline
Weighted WordVector+Wiki(Our Method) & 88.60\\ \hline
Weighted WordVector+TfIdf(Our Method) & 90.67\\ \hline
Composite Document Vector & \textbf{93.91}\\ \hline

\end{tabular}
\caption {Results on IMDB Movie Review Dataset}
\label{table:IMDB}
\end{table}

Table \ref{table:IMDB} summarizes the results obtained by others and by us on the IMDB movie review dataset. We have gone above the previous best~\cite{Le:14} by a margin of 1.33\% using discrimination weighting. The main contributor for improvement in results is our new document vector which overcomes the weaknesses of BOW and document vectors taken separately.

\begin{table}[h]
\centering
\small
\begin{tabular}{|l|l|l|l|}
\hline
\textbf{Method}                                                                 & \textbf{Weight} & \textbf{Accuracy(1)} & \textbf{Accuracy(2)} \\ \hline
\multirow{2}{*}{\begin{tabular}[c]{@{}l@{}}0-1 \\ Weighting\end{tabular}}       & 0               & 93.84                & 93.06                \\ \cline{2-4} 
                                                                                & 1               & \textbf{93.91}       & \textbf{93.18}       \\ \hline
\multirow{6}{*}{\begin{tabular}[c]{@{}l@{}}Graded idf \\ Weighting\end{tabular}} & 2               & \textbf{93.89}       & 93.17                \\ \cline{2-4} 
                                                                                & 2.5             & 93.87                & 93.16                \\ \cline{2-4} 
                                                                                & 2.8             & 93.86                & 93.16                \\ \cline{2-4} 
                                                                                & 3               & 93.86                & \textbf{93.22}       \\ \cline{2-4} 
                                                                                & 4               & 93.83                & 93.12                \\ \cline{2-4} 
                                                                                & 5               & 93.75                & 93.03                \\ \hline
\end{tabular}
\caption {Results on IMDB Movie Reviews using Various Weighting Techniques(Composite Document Vector);Accuracy(2) is when we exclude tf-idf features}
\label{table:graded_weighting_tfidf}
\end{table}
Tables \ref{table:graded_weighting_tfidf} demonstrate the effectiveness of our proposed graded weighting technique. Without \emph{tf-idf} features, our proposed method performs better in the graded idf weighting case and when we include \emph{tf-idf} features, 0-1 weighting perform better than idf graded technique and both perform better than the previous state-of-the-art. We see that with larger weights there is a decrease in accuracy and that is because we are now filtering out more words which are important while building document vector.
Table \ref{table:IMDB_rnnlm} is a further improvement in results once we incorporate predictions of RNNLM and Composite document vector model together(voting ensemble). Here, we first trained a \emph{RNNLM} and then obtained predictions on test reviews in terms of probability. We trained Linear SVM classifier using new Document Vectors and then obtained predictions on test reviews. We then merged these two predictions using a voting based approach to obtain final classification.

\begin {table}[H]
\centering
\small
\begin{tabular}{ | p{5.5cm} | p{1.2cm} | }
\hline
\textbf{Method} & \textbf{Accuracy} \\ \hline
Composite Document Vector & 93.91\\ \hline
Composite Document Vector + RNNLM (Our Method) & \textbf{94.19}\\ \hline
\end{tabular}
\caption {Results on IMDB Movie Review Dataset}
\label{table:IMDB_rnnlm}
\end{table}

Table \ref{table:amazon} presents result of experiment conducted on famous Amazon electronics review dataset~\cite{snapnets}. Our vector averaging method alone has beaten previous best by 3.3\%.
\begin {table}[H]
\centering
\small
\begin{tabular}{ | p{5.5cm} | p{1.2cm} | }
\hline
\textbf{Features} & \textbf{Accuracy} \\ \hline
\cite{Dredze:08} & 85.90\\ \hline
Max Entropy~\cite{Dredze:08} & 83.79\\ \hline
WordVector Averaging(Our Method) & 88.63\\ \hline
Composite Document Vector (Our Method) & 92.17\\ \hline
Composite Document Vector + RNNLM & \textbf{92.91}\\ \hline
\end{tabular}
\caption {Results on Amazon Electronics Review Dataset}
\label{table:amazon}
\end{table}

\begin {table}[h!]
\centering
\small
\begin{tabular}{ | p{3.6cm} | p{1.5cm} | p{1.5cm} | }
\hline
\textbf{Features} & \textbf{Accuracy(1)} & \textbf{Accuracy(2)} \\ \hline
WordVector Averaging & 78.0 & 79.62\\ \hline
WordVector+tf-idf & 90.73 & 89.52\\ \hline
WordVector+tf-idf without stop words & 91.14 & 89.97\\ \hline
Weighted WordVector & 89.71 & 85.90\\ \hline
Weighted WordVector+tfidf & \textbf{92.89} & \textbf{90.30}\\ \hline
\end{tabular}
\caption {Accuracies for Product Review and Movie Review Datasets.}
\label{table:hindi_ourmethods}
\end{table}

Table \ref{table:hindi_ourmethods} represents the results using five different techniques for feature set construction. We see that there is a slight improvement in accuracy on both datasets once we remove stop-words but the major breakthrough occurs once we used weighted averaging technique for construction of document vectors from word vectors.

\begin {table}[H]
\centering
\small
\begin{tabular}{ | p{3.3cm} | p{2.2cm} | p{1.15cm} | }
\hline
\textbf{Experiment} & \textbf{Features} & \textbf{Accuracy} \\ \hline
In-language with SVM \cite{Joshi:10} & tfidf & 78.14\\ \hline
MT Based with SVM \cite{Joshi:10} & tfidf & 65.96\\ \hline
Improved HindiSWN \cite{Bakliwal:12} & Adj. \& Adv. presence & 79.0\\ \hline
WordVector Averaging & word vector & 78.0\\ \hline
Word Vector Averaging & word vector+tfidf & 89.97\\ \hline
Weighted Word Vector with SVM (Our method) & tfidf+weighted word vector & \textbf{90.30}\\ \hline
\end{tabular}
\caption {Comparison of Approaches: Movie Review Dataset}
\label{table:hindi_movie}
\end{table}

\begin {table}[H]
\centering
\small
\begin{tabular}{ | p{3.3cm} | p{2.2cm} | p{1.15cm} | }
\hline
\textbf{Experiment} & \textbf{Features} & \textbf{Accuracy} \\ \hline
Subjective Lexicon~\cite{Bakliwal:12} & Simple Scoring & 79.03\\ \hline
Hindi-SWN Baseline (Arora et al., 2013) & Adj. \& Adv. presence & 69.30\\ \hline
Word Vector with SVM & word vector+tfidf & 91.14\\ \hline
Weighted Word Vector with SVM (Our method) & tfidf+weighted word vector & \textbf{92.89}\\ \hline
\end{tabular}
\caption {Comparison of Approaches: Product Review Dataset}
\label{table:hindi_product}
\end{table}
Table \ref{table:hindi_movie} and \ref{table:hindi_product} compares our best method with other methods which have performed well using techniques such as tf-idf, subjective lexicon, etc.

\section{Conclusion}
\label{sec:conclusion}
In this work we present an early experiment on the possibilities of distributional semantic models (word vectors) for low-resource, highly inflected languages such as Hindi.  What is interesting is that our word vector averaging method along with tf-idf results in improvements of accuracy compared to existing state-of-the art methods for sentiment analysis in Hindi (from 80.2\% to 90.3\% on IITB Movie Review Dataset). Also from Table \ref{table:3Datasets}, we can see that paragraph vector proposed by \cite{Le:14} doesn't perform well owing to the fact that the Hindi dataset just contains single sentences highlighting the weakness of this model. The size of the corpus is also small to learn paragraph vectors. Thus, our model overcomes these weaknesses with a better document representation.
We observe that pruning high-frequency stop words improves the accuracy by around 0.45\%. This is most likely  because such words tend to occur in most of the documents and don't contribute to sentiment. For example, the word {\dn EPSm}(Film) occurs in 139/252 documents in Movie Reviews(55.16\%) and has little effect on sentiment. Similarly words such as {\dn Es\388wAT\0}(Siddharth) occur in 2/252 documents in Movie Reviews(0.79\%). These words don't provide much information.

We also see that when number of features accumulate to a large number than there are few redundant features creating noise in the representation of the text. We tried to reduce this noise by using feature variance techniques. The large increase in accuracy(around 11\%) justifies our claim.

Before concluding, we return to the unexpectedly high improvement in accuracy achieved. One possibility we considered is that when the skip-grams are learned from the entire review corpus, it incorporates some knowledge of the test data.  But this seems unlikely since the difference in including this vs not including it, is not too significant.  The best explanation may be that the earlier methods, which were all in some sense based on a sentiWordnet, and at that one that was initially translated from English, were essentially very weak.  This is also clear in an analysis from
~\cite{Bakliwal:12}, which shows intern-annotator agreement on sentiment words are very poor (70\%) - i.e. about 30\% of these words have poor human agreement. Compared to this, the word vector model  
provides considerable power underlining the claim that distributional semantics is a topic worth exploring for Indian languages.

Our experiments on new dataset and existing datasets show that our method is competitive with existing methods including state-of-the-art. This new concept of document vectors can overcome the weaknesses of existing models which were either deficient in capturing syntactic or semantic properties of text. These models failed to incorporate contribution of each word while we have tapped this area and hence achieved state-of-the art results. The ensemble of RNNLM and Composite Document Vector has beaten state-of-the-art by a significant margin and has opened this area for future research. These models have the advantage that they don't require parsing at any step neither do they require a lot of heavy pre-processing. These tasks require a lot of extra effort and they slow the progress a lot.

\section{Future Work}
\label{sec:future_work}
Distributional semantics approaches remain relatively under-explored for Indian languages, and our results suggest that there may be substantial benefits to exploring these approaches for Indian languages.  While this work has focused on sentiment classification, it may also improve a range of tasks from verbal analogy tests to ontology learning, as has been reported for other languages.
For future work, we can explore various compositional models - a) weighted average - where weights are determined based on cosine distances in vector space;  b) weighted multiplicative models. Identifying morphological variants would be another direction to explore for better accuracy. With regard to sentiment analysis, the idea of aspect-based models (or part-based sentiment analysis), which looks into constituents in a document and classify their sentiment polarity separately, remains to be explored in Hindi.

In English, our Composite document vectors has led open a new area to look at where there can be many possible ensembles which may improve our work. Also, we could incorporate multiple word vectors here as well to distinguish between polysemous words. Another interesting and open area is to look at \emph{Region of Importance} in NLP where we filter out sentiment oriented sentences and phrases from a unfocused corpus which contains text from various domains. The code and parameters are available at \emph{github} for future research.

\bibliographystyle{icon}
\bibliography{icon2015}

\end{document}